\documentclass[conference,flushend]{iaria}
\usepackage{booktabs}
\usepackage{enumitem}
\usepackage{balance}
\usepackage[acronym]{glossaries}
\usepackage{caption}
\newacronym{ml}{ML}{Machine Learning}
\newacronym{tl}{TL}{Transfer Learning}
\newacronym{zsl}{ZSL}{Zero-Shot Learning}
\newacronym{gnn}{GNN}{Graph Neural Network}
\newacronym{gcn}{GCN}{Graph Convolutional Network}
\newacronym{gat}{GAT}{Graph Attention Network}
\newacronym{gin}{GIN}{Graph Isomorphism Network}
\newacronym{sage}{GraphSAGE}{Graph Sample and Aggregate}
\newacronym{ecnn}{DGCN}{Dynamic Graph Convolutional Network}
\newacronym{cnn}{CNN}{Convolutional Neural Network}
\newacronym{ann}{ANN}{Artificial Neural Network}
\newacronym{pf}{PF}{Power Flow}
\newacronym{opf}{OPF}{Optimal Power Flow}
\newacronym{se}{SE}{State Estimation}
\newacronym{nse}{NSE}{Neural State Estimation}
\newacronym{te}{TE}{Topology Estimation}
\newacronym{psse}{PSSE}{Power System State Estimation}
\newacronym{wls}{WLS}{Weighted Least Squares}
\newacronym{der}{DER}{Distributed Energy Resource}
\newacronym{drl}{DRL}{Deep Reinforcement Learning}
\newacronym{mse}{MSE}{Mean Squared Error}
\title{On Zero-shot Learning in Neural State Estimation of Power Distribution Systems}
\author{
  \IEEEauthorblockN{
    \orcidlink{0000-0002-5332-5760} Aleksandr Berezin\IEEEauthorrefmark{1}, \orcidlink{0000-0002-2018-1078} Stephan Balduin\IEEEauthorrefmark{1}, \orcidlink{0000-0003-2487-7475} Eric MSP Veith\IEEEauthorrefmark{1},\\ \orcidlink{0000-0001-5805-5408} Thomas Oberließen\IEEEauthorrefmark{2} and \orcidlink{0000-0001-6311-6113} Sebastian Peter\IEEEauthorrefmark{2}} 
  \IEEEauthorblockA{\IEEEauthorrefmark{1}OFFIS -- Institute for Information Technology, Oldenburg, Germany\\
  e-mail: {\tt$\lbrace$name.surname$\rbrace$@offis.de}}
  \IEEEauthorblockA{\IEEEauthorrefmark{2}Institute of Energy Systems, Energy Efficiency and Energy Economics\\
  Technical University of Dortmund, Germany}
}

\begin{document}
\maketitle

\begin{abstract}
This paper addresses the challenge of neural state estimation in power distribution systems. We identified a research gap in the current state of the art, which lies in the inability of models to adapt to changes in the power grid, such as loss of sensors and branch switching, in a zero-shot fashion. Based on the literature, we identified graph neural networks as the most promising class of models for this use case. Our experiments confirm their robustness to some grid changes and also show that a deeper network does not always perform better. We propose data augmentations to improve performance and conduct a comprehensive grid search of different model configurations for common zero-shot learning scenarios.
\end{abstract}


\section{Introduction}
\label{sec:introduction}

\Gls{psse} is the task of inferring the ``state'' of an electrical power grid from real-time data collected by various sensors distributed across the system. The ``state'' in this context generally refers to the voltage magnitudes and phase angles at each bus in the grid.

For many years, \gls{psse} was mainly performed for the transmission grids using simplifying assumptions such as near-DC power flow and computational methods with poor scalability \cite{wu1990power}. This is enabled by balanced operation with a relatively simple, predominantly linear topology of transmission grids, given their scale and structure.

On the contrary, distribution grids, which transport electricity from substations to end consumers, present distinct challenges. Their unbalanced nature, radial or weakly meshed topology, high R/X ratios, and cost inefficiency to achieve sufficient sensor coverage complicate the state estimation process. Initially designed with transmission systems in mind, conventional methods often struggle to provide accurate state estimation in these more complex, dynamic, and less predictable distribution systems \cite{wu1990power}.

However, with the proliferation of \glspl{der} and other complex consumers, grid operators are facing the necessity of performing \gls{psse} for distribution grids. Additionally, §14a of the German Energy Industry Act effectively requires operators to develop transparency in distribution grids in order to align consumption with production from renewable energy sources, which requires \gls{psse}.

In this paper, we begin by reviewing relevant prior work in Section~\ref{sec:related-work}, followed by a formal statement of our research question in Section~\ref{sec:rq}. Section~\ref{sec:methodology} details the methodology, including model selection, data preprocessing, and experimental setup. We present and analyze our results in Section~\ref{sec:eval} and discuss their implications. Finally, Section~\ref{sec:conclusion} summarizes our findings and suggests directions for future work.

\section{Related Work}
\label{sec:related-work}

The traditional and most widely-used approach for \gls{psse} is the \gls{wls} method \cite{wls}. This algorithm minimizes the sum of the squared differences between the observed and estimated measurements, with each term being weighed inversely proportionally to the square of the measurement error standard deviation. 

However, the \gls{wls} algorithm is computationally intensive. Its time complexity is generally considered to be $\mathcal{O}(N^3)$ in the number of buses $N$, assuming a dense system matrix \cite{wls}. This is due to the need for matrix inversions and solving linear equations. This complexity can become a limitation for large-scale power systems with thousands of buses, leading to significant computational burden and time constraints, especially when real-time or near-real-time estimations are required. Additionally, \gls{wls} assumes that all error distributions are Gaussian, a condition that may not always hold true in practice.

To overcome these limitations, an increasing number of publications instead use \glspl{ann} for \gls{psse}, a combination that is called \gls{nse}. \Glspl{ann} may be able to perform the calculation faster than iterative solutions and achieve a higher solution quality simultaneously \cite{mestav2018state}\cite{balduin2020evaluating}. However, like all \gls{ml} methods, the performance of \glspl{ann} is contingent on the quality and quantity of the available training data. Therefore, \gls{nse} approaches are usually valid only for the grid they have been trained on. Once the topology or characteristics of nodes change, the \gls{ann} needs to be retrained. This is known as the problem of \gls{tl}.

The most logical way to overcome this limitation is to use models that incorporate information about the graph topology into their calculations. Such models are known under an umbrella term \glspl{gnn}. Expectedly, recent years have seen a high volume of publications that propose utilizing \glspl{gnn} for \gls{nse} in various ways. To name a few examples:

\begin{itemize}
    \item \citeauthor{nse1} \cite{nse1} lays important groundwork in comparing different matrix representations of graphs within the \gls{gcn} model;
    \item \citeauthor{nse2} \cite{nse2} utilizes \glspl{gat} with a different graph representation of the power grid;
    \item \citeauthor{nse3} \cite{nse3} explore the possibility of utilizing temporal correlations in the datasets using recurrent \glspl{gcn}.
\end{itemize}

However, to our knowledge, none of these research projects specifically considered the problem of \gls{zsl} in \gls{psse}. The contribution of this work is in setting up multiple evaluation scenarios for \gls{zsl} and testing different configurations of \glspl{gnn} in them.

\section{Research question}
\label{sec:rq}

When discussing the ability of a model to generalize to different grid topologies, it is important to differentiate between \textit{homogeneous} and \textit{heterogeneous} modes of \gls{tl}. In general, homogeneous \gls{tl} mode means that the source and target data are in the same feature space, while in heterogeneous \gls{tl} mode, they are represented in different feature spaces.

In the context of power grids, this is the difference between two use cases. In the homogeneous case, the power grid remains the same, but some connections between its nodes appear or disappear due to changes in switch states or elements going in and out of service. In the heterogeneous case, the model trained on one grid is used to make predictions about a completely different grid \cite{Yang2021Power-grid}.

This distinction becomes very important in production environments. Integrating a model into the control system of a real grid naturally takes time, and training the model on that specific grid could be incorporated into this process without noticeably slowing it down. On the other hand, changes in grid topology due to switching can happen suddenly and unpredictably, and the model must adapt to them in real-time. 

There is also another way in which the data distribution can shift in the context of \gls{psse}: the observable subset of buses can change, which changes the amount of input data points available to the model. This can also be considered a form of homogeneous \gls{tl}.

A subset of \gls{tl} is \acrfull{zsl}. This scenario excludes the possibility of fine-tuning the model on the new distribution and evaluates its performance directly after the transfer. In this project, we specifically focus on \gls{zsl} because it is more representative of real-life situations where a model must make predictions immediately after a topology change without access to any training data for fine-tuning. In other words, the model should be \textit{robust} to distributional shifts.

Of course, in practice, a model can be fine-tuned to provide the best performance for the new topology. Still, until this process is complete, the previous version of the model has to substitute for it and provide good enough estimations, even if they are of lower quality.

The research question for this paper is which existing models in application to the \gls{psse} problem are robust to changes in the data distribution, specifically:

\begin{enumerate}[label=\Alph*]
    \item To the reduction of the subset of observable buses;
    \item To grid topology changes resulting from changing switch states;
    \item To transfer to a completely different power grid.
\end{enumerate}

\section{Methodology}
\label{sec:methodology}

\subsection{Model selection}
\label{sec:models}

The general question of model selection for \gls{nse} was addressed by us previously in \cite{modelcomparison}. The main conclusion from that paper was the selection of \glspl{gnn} as the most promising direction for further research. Now, we will perform a similar comparison study within the \gls{gnn} family. We are comparing four models using the implementations provided by PyTorch Geometric framework \cite{pyg}:

\begin{enumerate}
    \item \Acrfull{gcn} as proposed in \cite{gcn}
    \item \Acrfull{gat} as proposed in \cite{gat}
    \item \Acrfull{gin} as proposed in \cite{gin}
    \item \Acrfull{sage} as proposed in \cite{sage}
\end{enumerate}

\subsection{Graph representation of power systems}

A successful application of \gls{gnn} models naturally depends on how well the underlying data can be represented in the graph format. The first step is to represent buses in the grid as nodes of the graph and lines as its edges. In this project, we also represented transformers as edges without any additional parameters. For this to work, the voltage levels across the transformer must be normalized to avoid large voltage gradients.

It is also theoretically advantageous to use a weighted graph with line admittances as weights. Admittances are chosen because the graph Laplacian operator assumes higher edge weights to mean a higher correlation between nodes. This operator is, in turn, used in both the \gls{gnn} models and the feature propagation algorithm discussed in the next subsection. It should be noted that the models in question support neither complex-valued weights nor multidimensional weights, so we have to use the magnitude of the true complex impedance.

However, using admittance instead of impedance as edge weights becomes a problem for representing closed switches, which have zero impedance and, therefore, infinite admittance. This problem is solved by fusing buses connected by closed bus-to-bus switches into one bus. This is complicated because multiple closed switches are often connected to the same bus, so a naive approach of fusing adjacent buses in random order does not work. Instead, we use an iterative algorithm. Firstly, we build an auxiliary graph of just the closed bus-to-bus switches with buses as nodes and switches as edges. In this graph, nodes with a degree of one can be safely removed (fused with their adjacent buses). This will, in turn, lower the degree of the adjacent node. Eventually, every node will reach a degree of one and can be fused until every connected component of the auxiliary graph is fused into a single node.

\subsection{Data preprocessing}

\Gls{gnn} models are geometrically isodimensional, meaning that each output node must have a corresponding input node. This presents a problem for the \gls{psse} use case, where we lack input features for many, if not most, input nodes. The question, therefore, is: How do we initialize the missing features in the input data?

The solution we chose is the feature propagation algorithm from \cite{rossi2021fp}, which interpolates missing node-level features by solving a heat equation with known features as boundary conditions. This results in a smooth interpolation of features between known nodes and forms a starting point for the subsequent application of \glspl{gnn}.

\subsection{Datasets}

\begin{figure}
\centering
\includegraphics[width=0.3\textwidth]{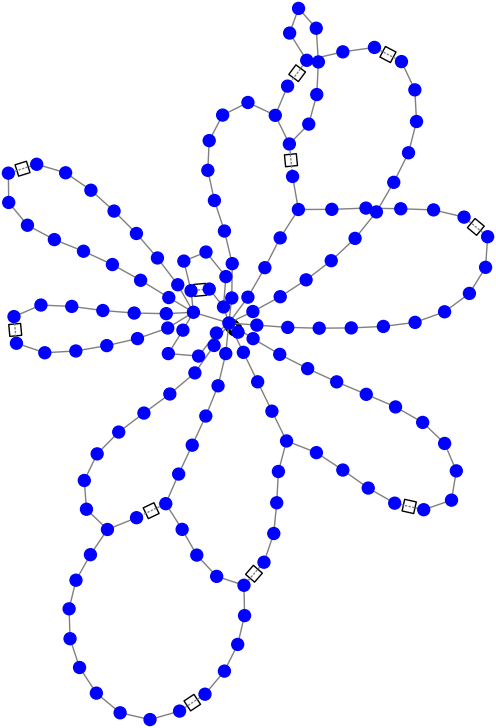}
\caption{A visualization of the SimBench 1-MV-urban–1-sw grid}
\label{fig:net}
\end{figure}

The main dataset used in this project is the SimBench \textit{1-MV-urban–1-sw}, a 147-node, 10 kV medium voltage grid \cite{meineckeSimBenchBenchmarkDataset2020} depicted in Figure \ref{fig:net}. It is composed of a grid model and a per-bus complex (active and reactive) power yearly time series. To calculate the resulting grid state, we performed a power flow calculation using SIMONA energy system simulation \cite{hiryAgentbasedDiscreteEventSimulation2022}. The resulting dataset comprises the base data and a year of complex voltage time series with a 15-minute temporal resolution. This dataset is hereafter called PQ.

Most grid branches in this model are of the open loop type, which means an open switch (depicted as a square) connects two separate branches. To simulate a realistic topology change, we made a line in one of the open loop branches inoperable, resembling a line fault, and closed the loop switch to resupply all nodes. Performing this operation on different branches resulted in multiple variations of the base grid topology. Afterward, we reran the simulation for each variation to obtain a topology change dataset, which is referenced hereafter as TC.

Unfortunately, the base dataset did not contain information about measurement devices. Therefore, we had to choose observable nodes randomly based on an observability level of 50\%, which we assume is realistic for distribution grids. This means that the state estimator has access to true voltage values for half of the grid buses.

An auxiliary dataset used in the heterogeneous \gls{zsl} experiments is based on the CIGRE medium voltage distribution network from Pandapower \cite{pandapower}. It is a much smaller grid with only 15 nodes, which allows us to study how the complexity of the grids affects the performance of \gls{zsl}. The voltage data for it is generated using the Midas simulation framework \cite{midas}. The shorthand name for this dataset is MV.

\subsection{Use cases}
\label{sec:uc}

Our experiments will be composed of three benchmarks that we call use cases. They correspond to the three subquestions of the main \nameref{sec:rq}.

In the first use case corresponding to subquestion A, we train the model on the grid with a baseline level of observability and then linearly reduce it from the baseline level to zero at testing time. Of course, the model performance decreases along with this reduction. The shorthand name of this use case is observability degradation (OD).

The second use case corresponds to subquestion B and tests \gls{zsl} for homogeneous topology changes. In it, we split the TC dataset in a 50:50 ratio, train the model on the first part, and evaluate on the second. We also evaluate another model trained on the PQ dataset on the TC testing subset to see if the model needs to observe the topology changes happening in order to be able to adapt to them at testing time, but our null hypothesis is that this is not the case. This scenario has the shorthand name TC1, and the former, where the model is trained on the TC dataset, is called TC2.

The third use case corresponds to subquestion C and covers the heterogeneous \gls{zsl} scenario. Here, we transfer the model between the PQ and MV datasets in both directions, that is, training on one and then testing on another. The scenario in which the model is trained on PQ and tested on MV has the shorthand name PQ2MV, and the other has MV2PQ.

\subsection{Experiment setup}

Let us now establish the full hyperparameter space for the models in question. It consists of the following dimensions:

\begin{itemize}
    \item Model, as listed in the \nameref{sec:models} subsection. Categorical parameter with four values.
    \item Number of layers in the model. Integer parameter that we limit to 10.
    \item Use of feature propagation (as opposed to initializing the missing features with zeros). Boolean parameter.
    \item Use of admittances as edge weights (as opposed to not using any edge weights). Boolean parameter.
\end{itemize}

Unfortunately, preliminary experiments have demonstrated that the hyperparameter space is not separable, meaning that a full grid search of the space is required. We performed this search for all model configurations and use cases and collected the \gls{mse} metric for each one. The results of this experiment are organized into an evaluation table where rows correspond to model configurations and columns are the following:

\begin{enumerate}
    \item ``model'' is the name of the model;
    \item ``layers'' is the number of layers in the model;
    \item ``fp'' is a binary parameter indicating whether feature propagation is used;
    \item ``adm'' is a binary parameter indicating whether admittance weights are used;
    \item ``mse'' is the value of \gls{mse} for the configuration defined by the above parameters.
\end{enumerate}

The full table is available in our repository in the ``results.csv'' file, and in the next section, we will use subsets of it as illustrations of results.

\section{Evaluation}
\label{sec:eval}

In this section, we will analyze the results of the full grid search, attempting to answer the following questions:

\begin{enumerate}
    \item Which model configurations perform best for each use case?
    \item How does model complexity affect performance?
    \item How do the data augmentations proposed in the \nameref{sec:methodology} section affect performance?
    \item How is performance on different tasks correlated?
\end{enumerate}

The answers to these questions will then be used to answer the main research questions from Section \ref{sec:rq}.

\subsection{Ranking model configurations}
\label{sec:ranking}

To interpret the numerical results listed in this section, it is useful to keep in mind the baseline value of \gls{mse} obtained by evaluating the models trained on the first half of the PQ dataset on the second half of the same dataset and taking the best result. This value is $0.32$. We can then broadly say that \gls{zsl} is possible in scenarios where the value of \gls{mse} after the topology change does not significantly exceed it.

\begin{table}
\caption{BEST CONFIGURATIONS FOR OBSERVABILITY DEGRADATION (OD)}
\label{res:od}
\centering
\begin{tabular}{lrrrr}
\toprule
model & layers & fp & adm & mse \\
\midrule
GraphSAGE & 3 & True & False & 0.86 \\
GraphSAGE & 2 & True & True & 0.87 \\
GraphSAGE & 3 & True & True & 0.87 \\
GraphSAGE & 2 & True & False & 0.88 \\
GCN & 3 & True & False & 0.90 \\
\bottomrule
\end{tabular}
\end{table}

The winning model for the first use case is \gls{sage} utilizing feature propagation. It also appears from Table \ref{res:od} that there is a sweet spot in model complexity of 2-3 layers.

\begin{table}
\caption{BEST CONFIGURATIONS FOR SWITCHING CHANGES (TC1)}
\label{res:tc1}
\centering
\begin{tabular}{lrrrr}
\toprule
model & layers & fp & adm & mse \\
\midrule
GraphSAGE & 3 & True & False & 0.31 \\
GraphSAGE & 1 & True & True & 0.33 \\
GAT & 1 & True & False & 0.33 \\
GraphSAGE & 2 & True & False & 0.34 \\
GraphSAGE & 3 & True & True & 0.34 \\
\bottomrule
\end{tabular}
\end{table}

\begin{table}
\caption{BEST CONFIGURATIONS FOR SWITCHING CHANGES (TC2)}
\label{res:tc2}
\centering
\begin{tabular}{lrrrr}
\toprule
model & layers & fp & adm & mse \\
\midrule
GraphSAGE & 1 & True & False & 0.32 \\
GraphSAGE & 1 & True & True & 0.32 \\
GAT & 3 & True & False & 0.32 \\
GAT & 1 & True & True & 0.32 \\
GAT & 1 & True & False & 0.32 \\
\bottomrule
\end{tabular}
\end{table}

In both scenarios of the second use case (Tables \ref{res:tc1} and \ref{res:tc2}), the model rating is similar, which suggests that the tasks themselves are similar as well. The winning models are \gls{sage} and \gls{gat}, also with the help of feature propagation and, curiously, in their shallowest versions, with single-layer models showing some of the best results. The main observation, however, is that the \gls{mse} values are identical to the baseline, meaning that homogeneous \gls{zsl} works very well.

\begin{table}
\caption{BEST CONFIGURATIONS FOR HETEROGENEOUS TRANSFER (PQ2MV)}
\label{res:pq2mv}
\centering
\begin{tabular}{lrrrr}
\toprule
model & layers & fp & adm & mse \\
\midrule
GCN & 1 & False & True & 0.30 \\
GCN & 2 & False & True & 0.87 \\
GCN & 4 & False & True & 0.97 \\
GAT & 2 & False & False & 1.06 \\
GraphSAGE & 2 & False & True & 1.26 \\
\bottomrule
\end{tabular}
\end{table}

\begin{table}
\caption{BEST CONFIGURATIONS FOR HETEROGENEOUS TRANSFER (MV2PQ)}
\label{res:mv2pq}
\centering
\begin{tabular}{lrrrr}
\toprule
model & layers & fp & adm & mse \\
\midrule
GAT & 6 & True & False & 0.62 \\
GAT & 2 & True & False & 0.64 \\
GCN & 2 & True & True & 0.67 \\
GIN & 1 & True & True & 0.69 \\
GAT & 2 & False & True & 0.78 \\
\bottomrule
\end{tabular}
\end{table}

In the third use case, we see a significant difference between the two scenarios. In the first scenario (Table \ref{res:pq2mv}), where the model is transferred from a larger to a smaller grid, the best-performing by a large margin is a single-layer \gls{gcn}, which is the simplest of all the compared models. A possible explanation is that there are only a few correlations that are reusable between grids, which the simple model can capture. Any more complex model picks up too many correlations that are specific to the grid it was trained on and then misapplies them. In the second scenario (Table \ref{res:mv2pq}), the results are mixed between complex and simple models, and we cannot come up with a sound theoretical interpretation of this result.

\subsection{Impact of model complexity}

\begin{figure*}
\centering
\includegraphics[width=0.9\textwidth]{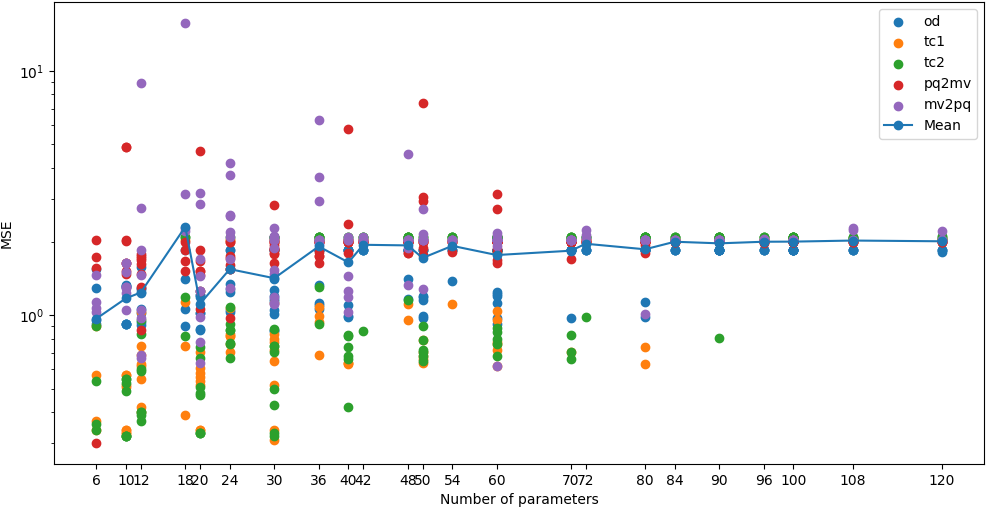}
\caption{Performance as a function of model complexity}
\label{fig:params}
\end{figure*}

In Figure \ref{fig:params}, we plot the performance of models against their trainable parameter counts. We also plot the average for all models of a given complexity: the ``Mean'' line on the graph.

Here, we can see a break point at about 84 parameters or 8 layers, starting from which the performance of models becomes much more consistent between use cases and configurations. The explanation for this effect is the over-smoothing phenomenon described in \cite{oversmoothing}. In short, \gls{gnn} layers of all architectures tend to act as low-pass filters, which effectively averages the output values over multiple iterations. Eventually, the model converges to an output where the values at all nodes of the graph are identical.

Since we are not interested in over-smoothed results, we can drop the model configurations that output them from further analysis. Therefore, the following sections will use the results table truncated to a maximum of 7 layers.

\subsection{Impact of data augmentations}

\begin{table}
\caption{IMPACT OF DATA AUGMENTATIONS}
\label{res:aug}
\centering
\begin{tabular}{rrrrrrr}
\toprule
fp & adm & od & tc1 & tc2 & pq2mv & mv2pq \\
\midrule
True & True & 1.33 & 1.17 & 1.13 & 2.19 & 1.80 \\
True & False & 1.25 & 1.03 & 1.06 & 1.99 & 2.55 \\
False & True & 1.51 & 1.24 & 1.25 & 1.93 & 2.00 \\
False & False & 1.49 & 1.23 & 1.10 & 1.98 & 2.09 \\
\bottomrule
\end{tabular}
\end{table}

In Table \ref{res:aug}, we average the performance values across all model configurations, leaving only the data augmentations as parameters. The results are unsurprising and follow the observations we made previously. Homogeneous scenarios benefit from feature propagation but are held back by admittance weights. In the heterogeneous scenarios, we once again see a split where MV2PQ benefits from admittance weights and PQ2MV does not. On average, the impact of the augmentations is not very significant.

\subsection{Correlation analysis}
\label{sec:corr}

\begin{figure}
\centering
\includegraphics[width=0.45\textwidth]{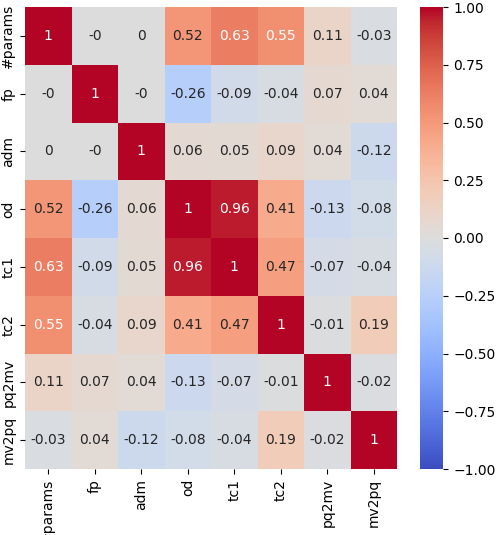}
\caption{Performance correlation between use cases}
\label{fig:corr}
\end{figure}

To explore the correlations between hyperparameters and use cases, we compute the Pearson correlation matrix between their associated performance values in Figure \ref{fig:corr}. Here, we also use the number of trainable parameters (``\#params'') instead of layers to compare model complexity more fairly. Note that since lower \gls{mse} means better performance, a positive correlation between a hyperparameter and performance is shown as negative and vice versa.

The analysis confirms the conclusions that we made previously: model complexity is detrimental to performance for all use cases except MV2PQ, and our proposed augmentations only marginally affect performance, with feature propagation being most useful in homogeneous scenarios.

\section{Limitations and future work}

During the evaluation stage of this research, it became evident that \gls{mse} alone does not convey enough information to confidently make conclusions about \gls{zsl} performance of the models. However, we could not find a better alternative in the literature.

The main problem is that \gls{mse} only shows us the instantaneous performance and does not account for the transfer process. An ideal metric $M$ for \gls{zsl} and \gls{tl} experiments would be a differential one that takes into account the magnitude of the change in the underlying grid $\Delta G$ and the performance change $\Delta P$, for example, $$M = \frac{\Delta P}{\Delta G}$$
However, an algorithm to compute $\Delta G$ is not trivial to develop. We hope to tackle this problem in our future work.

\section{Conclusion}
\label{sec:conclusion}

The findings of this paper can be summarized as follows:

\begin{enumerate}
    \item \Glspl{gnn} are very robust to homogeneous topology changes in the underlying power grid.
    \item Some \glspl{gnn} can perform well in a zero-shot transfer from a larger grid to a smaller one, but not in the other direction.
    \item Despite the conventional wisdom in the \gls{ml} community being ``Scale is all you need,'' scaling \glspl{gnn} up is not the best way to improve \gls{nse} performance. This is explained by the oversmoothing phenomenon \cite{oversmoothing}.
    \item Measuring the performance of \gls{nse} by \gls{mse} is not always helpful, especially in the context of \gls{zsl}. However, we could not find another commonly accepted metric. The development of such a metric appears to be a research gap at the moment.
\end{enumerate}

\subsubsection*{Acknowledgments}
This research is a part of project TRANSENSE, funded by the German Federal Ministry for Economic Affairs and Climate Action (FKZ~03EI6044A).

\subsubsection*{Availability of data and source code}
\label{source}
The source code for this project, along with the datasets, is openly available in the following repository:\\
\url{https://gitlab.com/transense/nse-tl-paper/tree/IARIA}

\balance
\printbibliography
\end{document}